%% file: nips_2020.tex
\documentclass{article}
\PassOptionsToPackage{numbers, compress}{natbib}

\usepackage[final]{nips_2020}
\usepackage{multirow}
\usepackage[utf8]{inputenc} 
\usepackage[T1]{fontenc}    
\usepackage[colorlinks,bookmarks=false]{hyperref}
\usepackage{url}            
\usepackage{booktabs}       
\usepackage{amsfonts}       
\usepackage{amsmath}
\usepackage{nicefrac}       
\usepackage{microtype}      
\usepackage{color}
\usepackage{graphicx}
\usepackage{subcaption}
\usepackage{xcolor}
\usepackage{enumitem}
\usepackage{bbm}
\usepackage{todonotes}

\input{math_cmds.tex}

\usepackage{floatrow}
\newfloatcommand{capbtabbox}{table}[][\FBwidth]
\usepackage{mathtools}
\usepackage[ruled,norelsize]{algorithm2e}
\newcommand{\ie}{i.e.,\xspace}
\definecolor{midnightgreen}{rgb}{0.0, 0.29, 0.33}

\newcommand{\abbr}{ELV\xspace}

\DontPrintSemicolon 

\newfloat{sbsfigure}{t}{ext}

\usepackage[utf8]{inputenc} 
\usepackage[T1]{fontenc}    
\usepackage{hyperref}       
\usepackage{url}            
\usepackage{booktabs}       
\usepackage{amsfonts}       
\usepackage{nicefrac}       
\usepackage{microtype}      
\usepackage{todonotes}
\usepackage{bbm}

\title{Towards Interpretable Natural Language Understanding with Explanations as Latent Variables}
\author{
Wangchunshu Zhou$^{1}$\thanks{\ Equal contribution, with order determined by rolling a dice. Work was done during internship at Mila.} ~~~ Jinyi Hu$^{2}$$^*$ ~~~ Hanlin Zhang$^{3}$$^*$ \\ ~~~ \bf Xiaodan Liang$^4$ ~~~ Maosong Sun$^2$ ~~~ Chenyan Xiong$^5$ ~~~ Jian Tang$^{6,7}$\\
$^1$ Beihang University $^2$ Tsinghua University $^3$ South China University of Technology \\
$^4$ Sun Yat-sen University
$^5$ Microsoft Research\\
$^6$ Mila-Québec AI Institute $^7$ HEC Montréal\\
{\tt zhouwangchunshu@buaa.edu.cn} \\
{\tt hujy17@mails.tsinghua.edu.cn}
{\tt sms@tsinghua.edu.cn} \\
{\tt \{hlzhang109, xdliang328\}@gmail.com} \\
{\tt chenyan.xiong@microsoft.com} \\
{\tt jian.tang@hec.ca}}
\begin{document}

\maketitle

\begin{abstract}

Generating natural language explanations has shown to be effective in not only offering interpretable explanations but also providing additional information and supervision for prediction. However, existing approaches usually require a large set of human annotated explanations for training while collecting a large set of explanations is not only time consuming but also expensive. In this paper, we develop a general framework for interpretable natural language understanding that requires only a small set of human annotated explanations for training. Our framework treats natural language explanations as latent variables that model the underlying reasoning process of a neural model. To optimize it, we develop a variational EM framework where an explanation generation module and an explanation-augmented prediction module are alternatively optimized and mutually enhance each other. Moreover, we further propose an explanation-based self-training method under this framework for semi-supervised learning. It alternates between assigning pseudo-labels to unlabeled data and generating new explanations to iteratively improve each other. Experiments on two natural language understanding tasks demonstrate that our framework can not only make effective predictions in both supervised and semi-supervised settings, but also generate good natural language explanations.\footnote{\ Code is available at \href{https://github.com/JamesHujy/ELV.git}{https://github.com/JamesHujy/ELV.git}}

\end{abstract}

\input{introduction}

\input{relatedwork.tex}
\input{method}

\input{experiments}

\section{Conclusion}

In this paper, we propose \abbr, a novel framework for training interpretable natural language understanding models with limited human annotated explanations. Our framework treats natural language explanations as latent variables that model the underlying reasoning process to enable interactions between explanation generation and explanation-based classification. Experimental results in both supervised and semi-supervised settings show that \abbr can not only make effective predictions but also generate meaningful explanations. In the future, we plan to apply our framework to other natural language understanding tasks. In addition, we plan to test the effectiveness of our framework on other pre-trained 
language models that are either stronger (e.g., XLNet~\citep{yang2019xlnet}, RoBERTa~\citep{liu2019roberta}, ALBERT~\citep{lan2019albert}, ELECTRA~\citep{clark2020electra}, etc.) or more computationally efficient. (e.g., DistilBERT~\citep{sanh2019distilbert}, BERT-of-Theseus~\citep{xu2020bert}, DeeBERT~\citep{xin2020deebert}, PABEE~\citep{zhou2020bert}, etc.)

\section*{Broader Impact}


On the other hand, such a system also brings potential risks depending on the quality of the generated natural language explanations. For example, the generated natural language could have certain biases, which have been reported in many natural language understanding systems~\cite{zhao2017men,costa2019analysis}. How to mitigate these risks will be our future work. Another potential risk is that the explanation generation model in our framework generates ad-hoc explanations that are not necessarily informative about how the model makes its predictions, since the model can come up with whatever explanation it thinks would pair with its predicted label. This is a common drawback for current explanation generation models. Our framework partially mitigates this problem since the generated explanations are in return used in the training process of the explanation-augmented classifier through the explanation retrieval process.

\section*{Acknowledgments}
This project is supported by the Natural Sciences and Engineering Research Council (NSERC) Discovery Grant, the
Canada CIFAR AI Chair Program, collaboration grants between Microsoft Research and Mila, Samsung, Amazon Faculty Research Award, Tencent AI Lab Rhino-Bird Gift Fund and a NRC Collaborative R\&D Project (AI4D-CORE-06). The authors would like to thank Meng Qu, Ziqi Wang, Cheng Lu, and the anonymous reviewers for their valuable feedback and insightful comments.

\bibliography{ref}{}
\bibliographystyle{unsrtnat}

\newpage
\appendix

\section{Examples of Explanations}

In this section we present several human annotated natural language explanations and explanationos generated by the explanation generation model in the \abbr framework in different datasets for better understanding of the proposed approach.

\subsection{TACRED}

\subsubsection{Human Annotated Explanations}
\begin{enumerate}[label=\alph*)]

\item
\textit{Although not a Playboy Playmate , she has appeared in nude pictorials with her Girls Next Door costars and fellow Hefner girlfriends Holly Madison and Kendra Baskett(OBJ), then known as Kendra Wilkinson(SUBJ).}

\textbf{Label: }\texttt{per:alternate\_names}

\textbf{Explanation: }The term "then known as" occurs between SUBJ and OBJ and there are no more than six words between SUBJ and OBJ
\item
\textit{Burke(SUBJ) 's mother Melissa Bell(OBJ) was a singer in the dance group Soul II Soul, which had hits in the 1980s and 1990s.}

\textbf{Label: }\texttt{per:parents}

\textbf{Explanation: }SUBJ and OBJ sandwich the phrase "'s mother" and there are no more than three words between SUBJ and OBJ
\item
\textit{Ellen Pompeo(OBJ) secretly married Chris Ivery(SUBJ) Congratulations to the newlyweds and let them live happily ever after !!}

\textbf{Label: }\texttt{per:spouse}

\textbf{Explanation: }There are no more than four words between SUBJ and OBJ and SUBJ and OBJ sandwich the phrase "secretly married"

\end{enumerate}
\subsubsection{Machine Generated Explanations}

\begin{enumerate}[label=\alph*)]

\item
\textit{Lomax shares a story about Almena Lomax, his mother and a newspaper owner and journalist(OBJ) in Los Angeles , taking her(SUBJ) family on the bus to Tuskegee, Ala., in 1961.}

\textbf{Label: }\texttt{per:title}

\textbf{Explanation: }the word "family" is right after SUBJ.

\item
\textit{What happened to their investments was of no interest to them, because they would already be paid , said Paul Hodgson(OBJ), senior research associate at the Corporate Library(SUBJ), a shareholder activist group.}

\textbf{Label: }\texttt{per:org:top\_members/employees}

\textbf{Explanation: }The word "senior research associate at" appears right before SUBJ.
\item

\textit{Iroquois passport dispute raises sovereignty issue The National Congress of American Indians(SUBJ), based in Washington, DC(OBJ), has advocated on behalf of the lacrosse team, urging British officials to allow the members entry into England on their Iroquois-issued passports.}

\textbf{Label: }\texttt{org:country\_of\_headquarterse}

\textbf{Explanation: }The word "based in" appears right before OBJ.

\end{enumerate}

\subsection{SemEval}
\subsubsection{Human Annotated Explanations}
\begin{enumerate}[label=\alph*)]
\item
\textit{Morton's SUBJ-O is the most common cause of localized OBJ-O in the third interspace and these diagnostic tests produce good indications of the condition.}

\textbf{Label: }\texttt{Cause-Effect(e1,e2)}

\textbf{Explanation: }Between SUBJ and OBJ the term "is the most common cause of" appears and SUBJ precedes OBJ
\item
\textit{The frontal SUBJ-O is a part of the OBJ-O that maintains very close ties with the limbic system.}

\textbf{Label: }\texttt{Component-Whole(e1,e2)}

\textbf{Explanation: }Between SUBJ and OBJ the term "is a part of the" appears and SUBJ precedes OBJ.
\item
\textit{Out current Secretary is gathering SUBJ-O from past OBJ-O and committee chairs.}

\textbf{Label: }\texttt{Entity-Origin(e1,e2)}

\textbf{Explanation: }The phrase "from past" links SUBJ and OBJ and there are no more than three words between SUBJ and OBJ and OBJ follows SUBJ.

\end{enumerate}

\subsubsection{Machine Generated Explanations}
\begin{enumerate}[label=\alph*)]
\item
\textit{SUBJ-O caused OBJ-O at the Charlotte Douglas International Airport Monday morning.}

\textbf{Label: } \texttt{Cause-Effect(e1,e2)}

\textbf{Explanation: }There is only one word "caused" between subj and obj and obj follows subj.

\item
\textit{The base in which it sits hides the damage which occurred when the SUBJ-O was removed from its initial OBJ-O on the Long Tan battlefield.}

\textbf{Label: }\texttt{Entity-Origin(e1,e2)}

\textbf{Explanation: }Between subj and obj the phrase "was removed into" occurs and there are no more than four words between SUBJ and OBJ and SUBJ precedes OBJ

\item
\textit{A hinge SUBJ-O attaches a OBJ-O pivotally to a base of an electronic device and has a pivoting leaf and a stationary leaf.}

\textbf{Label: }\texttt{Component-Whole(e1,e2)}

\textbf{Explanation: } The phrase "attaches a" between SUBJ and OBJ and OBJ follows SUBJ

\end{enumerate}

\subsection{Restaurant}
\subsubsection{Human Annotated Explanations}
\begin{enumerate}[label=\alph*)]
\item
\textit{We had great desserts (including the best cannoli I've ever had) and then they offered an after dinner drink, on the house. (Term: cannoli)}

\textbf{Label: }\texttt{positive}

\textbf{Explanation: }The word "best" directly precedes the term.
\item
\textit{All the desserts the group tried got favorable reviews. (Term: desserts)}

\textbf{Label: }\texttt{positive}

\textbf{Explanation: }The string "favorable" appears no more than 5 words after the term.
\item
\textit{The most annoying thing, though, is the fact that the servers seem to be trained to drive revenue. (Term: servers)}

\textbf{Label: }\texttt{negative}

\textbf{Explanation: }The word "annoying" occurs before the term.
\end{enumerate}

\subsubsection{Machine Generated Explanations}
\begin{enumerate}[label=\alph*)]
\item
\textit{This little place is wonderfully warm welcoming. (Term: place)}

\textbf{Label: }\texttt{positive}

\textbf{Explanation: }The terem is followed by "wonderful".
\item
\textit{The falafal was rather over cooked and dried but the chicken was fine. (Term: chicken)}

\textbf{Label: }\texttt{positive}

\textbf{Explanation:} The word "fine" occurs within 3 words after the term.
\item
\textit{Service was awful - mostly because staff were overwhelmed on a Saturday night. (Term: staff)}

\textbf{Label: }\texttt{negative}

\textbf{Explanation: }The word "unbearable" occurs within three words after the term. 

\end{enumerate}
\subsection{Laptop}
\subsubsection{Human Annotated Explanations}
\begin{enumerate}[label=\alph*)]
\item
\textit{The DVD drive randomly pops open when it is in my backpack as well, which is annoying. (Term: DVD drive)} 

\textbf{Label: }\texttt{negative}

\textbf{Explanation: }The string "annoying" occurs after the term
\item
\textit{The Apple team also assists you very nicely when choosing which computer is right for you. (Term: Apple team)}

\textbf{Label: }\texttt{positive}

\textbf{Explanation: }The string "very nicely" occurs after the term by no more than 6 words.
\item
\textit{The design is awesome, quality is unprecedented. (Term: design)}

\textbf{Label: }\texttt{positive}

\textbf{Explanation: }The word "awesome" is within 2 words after the term.
\end{enumerate}
\subsubsection{Machine Generated Explanations}
\begin{enumerate}[label=\alph*)]
\item

\textit{I ordered my 2012 mac mini after being disappointed with spec of the new 27 Imacs. (Term: spec)}

\textbf{Label: }\texttt{negative}

\textbf{Explanation: }The word "disappointed" occurs within 3 words before the term.
\item
\textit{I found the mini to be exceptionally easy to set up. (Term: set up)}

\textbf{Label: }\texttt{positive}

\textbf{Explanation: }The phrase "exceptionally easy" occurs within 3 words before the term.
\item
\textit{However, there are MAJOR issues with the touchpad which render the device nearly useless. (Term: tourchpad)}

\textbf{Label: }\texttt{negative}

\textbf{Explanation: }The phrase "nearly useless" occurs within 3 words after the term.

\end{enumerate}
\section{Human Evaluation Details}
For human evaluation, we random sample 100 examples in the test set of the Restaurant dataset and use ELV to predict the labels of the selected examples. Then we use different compared models to generate explanations of the prediction results with both the input sentences and the predicted labels. Afterward, we invite 5 graduate students with enough English proficiency to score the explanations. The annotation scenarios include the explanantions’ informativeness(Info.), correctness (Corr.), and consistency (Cons.)  with respect to the model prediction. Specifically, the informativeness measures to what extent the generated explanation is helpful to understand the model's prediction output. The correctness measures whether the explanation is factually correct (e.g. word ``good'' before the terms leads to positive label while word ``annoying'' is negative). The consistency refers to whether the explanation is consistent with the input sentence (i.e. the description in the explanation is true w.r.t the input sentence).

\begin{table}[]
    \centering
    \resizebox{0.8\textwidth}{!}{
 \begin{tabular}{lccccc}
 \toprule
 Fraction of labeled data used & 60\% & 70\% & 80\% & 90\% & 100\% \\
 \midrule
  \abbr (ours) & 75.5 & - & - & - & - \\
\midrule
 BERT-base & 74.6 & 75.0 & 75.3 & 75.4 & 75.4 \\
 \bottomrule
 \end{tabular}
}{
 \caption{Results (Macro-F1) on ASC datasets in supervised setting with different fraction of labeled data used.}
 \label{tab:appendix}
}
\end{table}

\section{Addition Analysis Experiments}

In this section, we report additional experimental results comparing the performance of \abbr with 60\% of labeled data and the performance with the BERT-base model with 60\%, 70\%, 80\%, 90\%, and 100\% of labeled data to investigate to what extent human annotated explanations can replace human labeled examples. The result is shown in Table \ref{tab:appendix}. We find that \abbr can achieve or even exceed the performance of a BERT-base model trained with much more labeled data. This confirms that \abbr can effectively leverage human annotated explanations as additional information.

\end{document}

%% file: math_cmds.tex
\usepackage{amsmath,amsfonts,bm}

\DeclareMathAlphabet{\mathsfit}{\encodingdefault}{\sfdefault}{m}{sl}
\SetMathAlphabet{\mathsfit}{bold}{\encodingdefault}{\sfdefault}{bx}{n}

%% file: introduction.tex
\section{Introduction}

Building interpretable systems for natural language understanding is critical in various domains such as healthcare and finance. One direction is generating natural language explanations for prediction \cite{hancock-etal-2018-training,rajani-etal-2019-explain,camburu2018snli,rajani2019explain}, which has been shown to be effective as they can not only offer interpretable explanations for back-box prediction but also provide additional information and supervision for prediction~\cite{srivastava-etal-2017-joint,zhou2020nero,Wang2020Learning}. For example, given a sentence ``\textit{The only thing more wonderful than the food is the service.}'', a human annotator may write an explanation like ``\textit{Positive, because the word `wonderful' occurs within three words before the term food.}'', which is much more informative than the label ``\textit{positive}'' as it explains how the decision was made. Moreover, the explanation can serve as an implicit logic rule that can be generalized to other instances like ``\textit{The food is wonderful, I really enjoyed it.}''.

There are recent works~\cite{camburu2018snli,rajani2019explain} that study generating natural language explanations for predictions and/or leverage them as additional features for prediction. For example, \citet{camburu2018snli} learns a language model to generate natural language explanations for the task of natural language inference by training on a corpus with annotated human explanations. \citet{rajani-etal-2019-explain} proposes a two-stage framework for commonsense reasoning which first trains a natural language explanation model and then further trains a prediction model with the generated explanations as additional information. These approaches achieve promising performance in terms of both predictive results and explainability. However, a large number of labeled examples with human explanations are required, which is expensive and sometimes impossible to obtain. Therefore, we are looking for an approach that makes effective prediction, offers good explainability, while only requires a limited number of human explanations for training. 

In this paper, we propose such an approach. We start from the intuition that the explanation-augmented prediction model is able to provide informative feedback for generating meaningful natural language explanations.  Therefore, different from existing work which trains the explanation generation model and the explanation-augmented prediction model in separate stages, we propose to \textbf{jointly} train the two models. Specifically, taking the task of text classification as an example, we propose a principled probabilistic framework for text classification, where natural language \textbf{E}xplanations are treated as \textbf{L}atent \textbf{V}ariables (\abbr). Variational EM~\cite{palmer2006variational} is used for the optimization, and only a set of human explanations are required for guiding the explanation generation process. In the E-step, the explanation generation model is trained to approximate the ground truth explanations (for instances with annotated explanations) or guided by the explanation-augmentation module through posterior inference (for instances without annotated explanations); in the M-step, the explanation-augmented prediction model is trained with high-quality explanations sampled from the explanation generation model. The two modules mutually enhance each other. As human explanations can serve as implicit logic rules, they can be used for labeling unlabeled data. Therefore, we further extend our \abbr framework to an \textbf{E}xplantion-based \textbf{S}elf-\textbf{T}raining (ELV-EST) model for leveraging a large number of unlabeled data in the semi-supervised setting.

To summarize, in this paper we make the following contributions:
\begin{itemize}[leftmargin=*]
\item We propose a principled probabilistic framework called \abbr for text classification, in which natural language explanation is treated as a latent variable. It jointly trains an explanation generator and an explanation-augmented prediction model. Only a few annotated natural language explanations are required to guide the natural language generation process. 
\item We further extend \abbr for semi-supervised learning (the ELV-EST model), which leverages natural language explanations as implicit logic rules to label unlabeled data.
\item We conduct extensive experiments on two tasks: relation extraction and sentiment analysis. Experimental results prove the effectiveness of our proposed approach in terms of both prediction and explainability in both supervised and semi-supervised settings.
\end{itemize}

%% file: relatedwork.tex
\section{Related Work}

Natural language (NL) explanations have been proved very useful for both model explanations and prediction in a variety of tasks recently~\cite{srivastava2017joint,murty2020expbert,camburu2018snli,rajani2019explain,rajani2019explain}. Some early work ~\cite{goldwasser2014learning,srivastava2017joint,murty2020expbert} exploited NL explanation as additional features for prediction. For example, \citet{srivastava2017joint} converted NL explanations into classifier features to train text classification models. \citet{fidler2017teaching} used natural language explanations to assist in supervising an image captioning model. Very recently, \citet{murty2020expbert} proposed ExpBERT to directly incorporate NL explanations with BERT. However, most of these work require the explanations to be available in both training and testing instances, which is not realistic as annotating the explanation of a huge amount of instances is very time consuming and expensive. Moreover, the prediction becomes much easier once the explanations are given in the testing data.

There is some recent work that studied training a natural language explanation model and then used the generated explanations for prediction. For example, ~\citet{camburu2018snli} and \citet{rajani2019explain} proposed to separately train a model to generate NL explanations and a classification model that takes the generated explanations as additional input. Their approaches have shown very promising for improving the interpretability of classification models and increasing the prediction performance with explanations as additional features. However, their approaches require a large number of human annotated NL explanations to train the explanation generation model. Moreover, these approaches fail to model the interaction between generating NL explanations and exploiting NL explanations for prediction. As a result, there is no guarantee that the generated explanations reflect the decision-making process of the prediction model or beneficial to the prediction model. As reported by \citet{camburu2018snli} that interpretability comes at the cost of loss in performance. In this paper, we propose a principled probabilistic framework with explanations as latent variables to minimize the number of training instances with explanations by jointly training the natural language explanation module and the explanation-augmented prediction module. 

Another relevant direction is treating natural language explanations as additional supervisions for semi-supervised learning instead of as additional features~\cite{hancock2018training,Wang2020Learning}. For example,  \citet{hancock2018training} utilized a semantic parser to parse the NL explanations into logical forms (\ie ``labeling function''). The labeling functions are then employed to match the unlabeled examples either hardly~\cite{hancock2018training} or softly~\cite{Wang2020Learning} to generate pseudo-labeled datasets used for training models. However, these approaches require the explanations to be annotated in a form that can be accurately parsed by a semantic parser to form labeling functions, which may not be possible for many NLP applications. In our semi-supervised framework, semantic parsing is not required, and natural language explanations are interpreted with distributed representation obtained by pre-trained language models for labeling unlabeled data. 

%% file: method.tex
\section{Methodology}

\begin{figure}
    
    \centering
    \resizebox{0.9\textwidth}{!}{
    \includegraphics[width=\columnwidth]{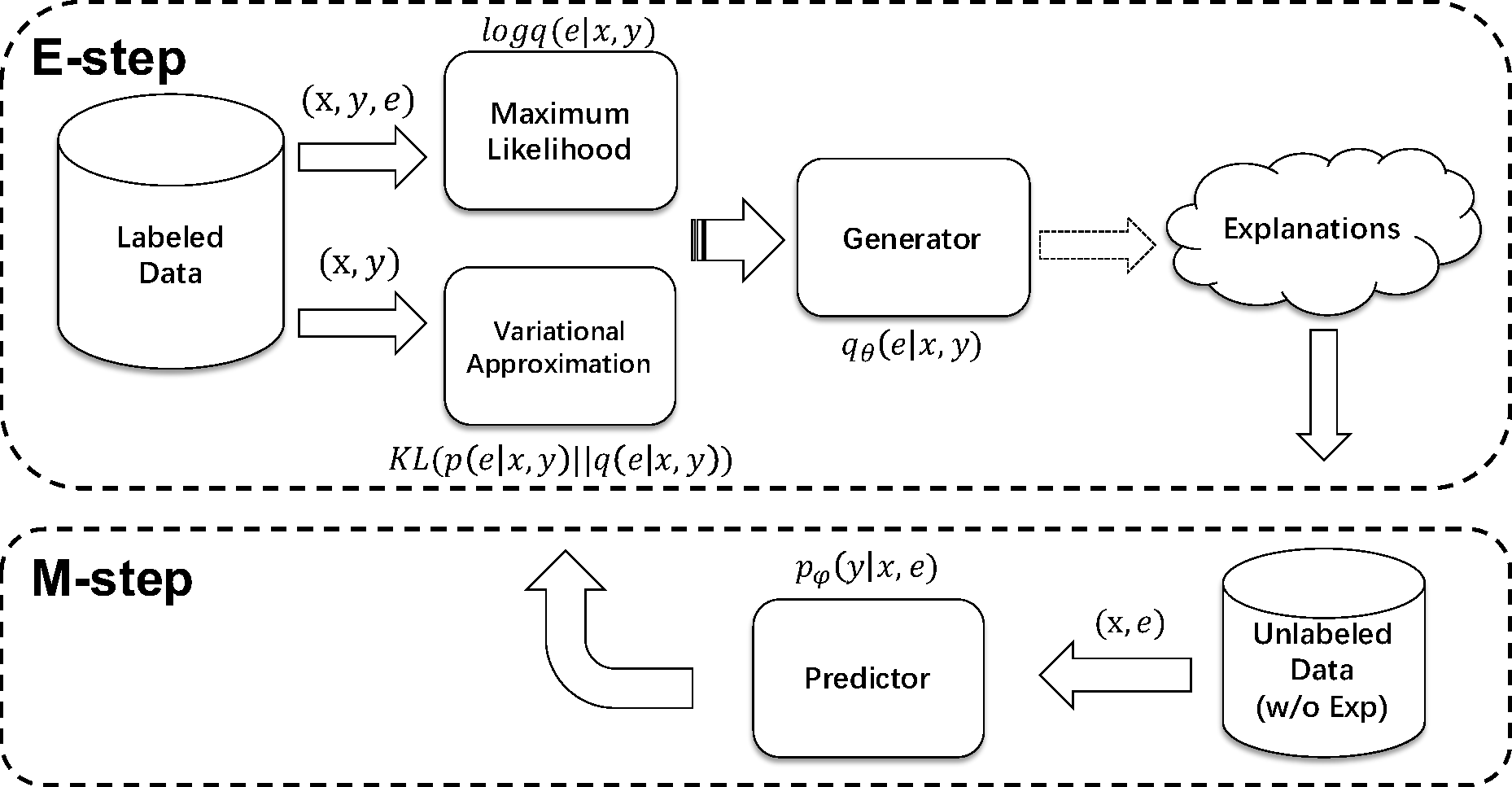}
    }
    \caption{
    Overview of \abbr.
    During E-step, we train our generator $p(e|x,y)$ to generate explanations given labeled data. For labeled data with annotated explanations (i.e. $\mathcal{D}_E$), we maximize the likelihood of the ground truth explanations. For labeled data without explanations (i.e. $\mathcal{D}_L$), we minimize the KL divergence between the variational distribution $q_\theta(e|x, y)$ and the ground truth posterior $p(e|x, y)$, which is calculated with the help of the prediction model. 
    During M-step, we use the explanation generated in E-step to train the predictor $p(y|x,e)$ with MLE. 
    }
    \label{fig:framework}
    \vspace{5pt}
\end{figure}{}

We first discuss the definition of the problem. Given an input sentence $x$, we aim to predict its label $y$ and generate a natural language (NL) explanation $e$ describing why $x$ is classified as $y$. Specifically, given a few amount of training example with NL explanation annotation $\mathcal{D}_E = \{ \left(x_1, y_1, e_1, \right), ... , \left(x_n, y_n, e_n, \right) \}$ and a relatively large set of labeled examples $\mathcal{D}_L = \{ \left(x_{n+1}, y_{n+1} \right), ... , \left(x_m, y_m\right) \}$, our goal is to learn: 1) An explanation generation model $E_\theta$ that parametrize $q(e|x,y)$, which takes a labeled example $(x,y)$ as input and generates a corresponding natural language explanation $e$, and 2) an explanation-augmented prediction model $M_\phi$, which parametrize $p(y|x,e)$ and takes an unlabeled example $x$ and NL explanations (as implicit rules) $E$ to assign a label $y$ to $x$.

\subsection{Natural Language Explanation as Latent Variables}

Given labeled data $(x,y)$, we treat the nature language explanation $e$ as a latent variable. For training, we aim to optimize the evidence lower bound (ELBO) of $\log p(y|x)$, which can be formulated as:
\begin{align}
    \log p(y|x) &= \log \int p(e, y|x) de =\log \int q_\theta(e|x,y)\frac{p(e, y|x)}{q_\theta(e|x,y)} de \\
                &\ge \mathcal{L}(\theta, \phi) = E_{q_\theta(e|x,y)} \log\frac{p(e, y|x)}{q_\theta(e|x,y)} = E_{q_\theta(e|x,y)} \log\frac{p_\phi(y|e, x)p(e|x)}{q_\theta(e|x,y)},
\end{align}
where $q_\theta(e|x,y)$ is the variational distribution of posterior distribution $p(e|x,y)$, $p(e|x)$ is the prior distribution of explanation $e$ for instance $x$, and $p_{\phi}(y|e,x)$ is the explanation-augmented prediction model. 

Due to the large search space of natural language explanation $e$, instead of using reprametrization trick used in variational autoencoder~\cite{kingma2013auto}, here we use the variational-EM algorithm for optimization. The overview of \abbr is illuatrated in Figure \ref{fig:framework}. Note that in our training data, the explanations of a few number of labeled examples (i.e., $\mathcal{D}_E$) are given. Therefore, we first initialize the explanation generation model $E_\theta=q_{\theta}(e|x,y)$ and the prediction model $M_\phi=p_{\phi}(y|e,x)$  by training on $\mathcal{D}_E$ with maximum likelihood estimation (MLE). Then, we update the aforementioned models by maximizing the log-likelihood $\log p(y|x)$ using $\mathcal{D}_L \cup \mathcal{D}_E$ with variational-EM. In the variational E-step, we train the explanation generator to minimize the KL divergence between $q_\theta(e|x,y)$ and $p(e|x,y)$, which will be detailed in Section 3.3. In the M-step, we fix $\theta$ and $p(e|x)$ and update the parameters $\phi$ of the prediction model to maximize the log-likelihood $\log p(y|x)$.

\subsection{E-step: Explanation Generation model}

As the core component of our variational EM framework, the explanation generation model is expected to generate ``soft'' logic rules in the form of natural language explanations. However, training a seq2seq model to generate explanations of high quality from scratch is very challenging. Motivated by the recent finding that pretrained language models encode various types of factual knowledge and commonsense knowledge in their parameters~\citep{bouraoui2019inducing,petroni2019language,roberts2020much}, we employ UniLM~\citep{dong2019unified}---a unified pre-trained language generation model that achieved state-of-the-art performance on many text generation tasks---as the explanation generation model $E_\theta$ in our framework. Specifically, the explanation generation model takes as input the concatenation of input sentence $x$ and the text description of its corresponding label $y$ to generate an explanation, which explains the label decision in natural language and can be treated as an implicit logic rule that can generalize to other examples.

Note that in the training data, only a small set of labeled examples are provided with explanations. Therefore, in the variational E-step, for labeled data $(x,y)$ without explanations (i.e. $\mathcal{D}_L$), we are trying to use the variational distribution  $q_{\theta}(e|x,y)$ to approximate the ground truth posterior $p(e|x,y)$, which can be calculated as
\begin{equation}
p(e|x,y) \sim p_{\phi}(y|x,e)p(e|x)    
\end{equation}
where $p_{\phi}(y|x,e)$ is parameterized by the prediction model and provides feedback for generating meaningful natural language explanations. We will introduce the detailed parametrization of $p(e|x)$ and $p_{\phi}(y|x,e)$ in the M-step.

For labeled data with explanations (i.e. $\mathcal{D}_E$), we just need to maximize the likelihood of the ground truth explanations. Therefore, the overall objective function of E-step can be summarized as:
\begin{equation}
 O = \sum_{(x,y)\in \mathcal{D}_E} \log q(e|x,y) + \sum_{(x,y)\in \mathcal{D}_L} \mathrm{KL}(q(e|x,y)\|p(e|x,y)) 
\end{equation}

\subsection{M-step: Explanation-Augmented Prediction model}
\label{mstep}

During M-step, the explanation-augmented prediction model is trained to predict the label of input sentence $x$ with the explanation $e$ generated from the variational distribution $q(e|x,y)$. 

However, note that the label $y$ is not available during testing, and the explanations for the unlabeled $x$ can only be generated from the prior distribution $p(e|x)$. Therefore, there are some discrepancies between the distributions of the explanations for labeled data in the training stage and those for unlabeled data in the testing stage since generating a natural language explanation without conditioning on a label is harder. To mitigate this issue, in the prediction model, besides sampling an explanation from the variational distribution, we also add a set of explanations from $p(e|x)$, which retrieves a set of explanations from similar sentences. 

Specifically, given an input sentence $x$ and the set of labeled and pseudo-labeled data consists of $(x',e',y')$, we retrieve $N$ explanations $\mathcal{E} \coloneqq \{{e'_i}\}^N_{i=1}$ of which the corresponding sentences $x'$ are the most similar to the input sentence $x$, measured by the cosine similarity between the embedding of $x$ and each $x'$ from $\mathcal{D}_E$ under SentenceBERT~\citep{reimers-gurevych-2019-sentence}, a pretrained sentence embedding model. Note that we do not directly use a seq2seq model to parametrize $p(e|x)$ because we find generating explanations without predicted labels often results in irrelevant and even misleading explanations.
\begin{algorithm}[h]
\caption{Explanation-based Self-Training (ELV-EST)}
\SetAlgoLined
\KwIn{$\mathcal{D}_E = \{ \left(x_1, y_1, e_1, \right), ... , \left(x_n, y_n, e_n, \right) \}$, $\mathcal{D}_L = \{ \left(x_{n+1}, y_{n+1} \right), ... , \left(x_m, y_m\right) \}$, unlabeled data $\mathcal{D}_U = \{ x_{m+1}, \dots, x_{N} \}$, Confidence threshold T}
\KwOut{$E_{\theta}(e|x,y)$,$M_{\phi}(y|x,E)$}
initialize $E_{\theta}$ and $M_{\phi}$ with $\mathcal{D}_E \cup \mathcal{D}_L$ using \abbr\;
\Repeat{Convergence or $\mathcal{D}_U = \emptyset$}{
    \For{each $x_i \in \mathcal{D}_U$}{\If{$\max_{y}M_{\phi}(y|x,E) > T$}{Assign pseudo-label $y_i$ to $x_i$ and generate explanation $e_i$ with $E_\theta$ \\ Update $\mathcal{D}_L$ = $\mathcal{D}_L \cup (x_i,y_i)$\\ Update $\mathcal{D}_E$ = $\mathcal{D}_E \cup (x_i,y_i,e_i)$ (for explanation retrieval)\\ Update  $\mathcal{D}_U = \mathcal{D}_U \setminus x_i$}}\
    Train $E_{\theta}$ and $M_\phi$ on $\mathcal{D}_E \cup \mathcal{D}_L$ with \abbr \;
}
\label{algo:main}
\end{algorithm}
\vspace{-0cm}

Let $\mathcal{E}=\{e_1,...,e_n\}$ denotes all the explanations of $x$. For each $e_i \in \mathcal{E}$, we feed the explanation $e_i$ and the input sentence $x$, separated by a \texttt{[SEP]}token, to BERT~\citep{devlin-etal-2019-bert} and use the vector at the \texttt{[CLS]} token to represent the interactions between $x$ and $e_i$ as a 768-dimensional feature vector:
\begin{equation}
    \mathcal{I}(x,e_i) = \text{BERT}([\texttt{[CLS]};x;\texttt{[SEP]};e_i])
\end{equation}
Our final classifier takes the concatenation of these vectors and outputs the final prediction as:
\begin{equation}
\label{eq:M}
    M_{\phi}(y|x,\mathcal{E}) = \text{MLP}\left[\text{Average}(\mathcal{I}(x,e_1);\mathcal{I}(x,e_2); ... ; \mathcal{I}(x,e_n))\right]
\end{equation}
At test time,  for each unlabeled $x$, we first use $p(e|x)$ to retrieve a set of explanations and then predict a label with the explanation-augmented prediction model. Afterward, we can further employ the explanation generation model to generate an NL explanation to explain the prediction decision based on both the input sentence and the predicted label.

To summarize, by alternating between E-step and M-step where $q_\theta(e|x,y)$ and $p_\phi(y|e, x)$ are optimized respectively, the explanation generation model $E_{\theta}$ and the explanation-augmented prediction model $M_{\phi}$ are jointly optimized and mutually enhanced. Next, we describe how our framework can be applied to the semi-supervised setting where both human-annotated explanations and ground-truth labels are limited.

\subsection{Explanation-based Self-Training}

As natural language explanations can serve as implicit logic rules, which can generalize to new data and help assign pseudo-labels to unlabeled data. Therefore, we extend the \abbr to the semi-supervised learning setting and propose an \textbf{E}xplanation-based \textbf{S}elf-\textbf{T}raining (ELV-EST) algorithm. In this setting, we only have limited labeled examples but abundant unlabeled data $\mathcal{D}_U = \{ x_{m+1}, \dots, x_{N} \}$.

As illustrated in Algorithm \ref{algo:main}, we first use \abbr to initialize $E_\theta$ and $M_\phi$ with the limited labeled corpus $\mathcal{D}_E \cup \mathcal{D}_L$. Afterward, we iteratively use $M_\phi$ to assign pseudo-labels to unlabeled examples in $\mathcal{D}_U$ to extend the labeled data $\mathcal{D}_L$. We then use \abbr to jointly train $E_\theta$ and $M_\phi$ with the augmented labeled dataset. At the same time, we also employ $E_\theta$ to generate new explanations with unlabeled examples and their pseudo-labels. In this way, we can harvest massive pseudo-labels and pseudo-explanations with unlabeled examples. The pseudo-labeled examples can be used to improve the models while also enable us to generate more NL explanations. In return, the newly generated explanations can not only improve the explanation generation model but also serve as implicit rules that help the prediction model assign more accurate pseudo-labels in the next iteration.

The proposed ELV-EST approach is different from the conventional self-training method in two perspectives. First, in addition to predicting pseudo-labels for unlabeled data, our method also discovers implicit logic rules in the form of natural language explanations, which in return helps the prediction model to better assign noisy labels to the unlabeled data. Second, our approach can produce explainable predictions with $E_\theta$. 
Compared to recent works \citep{Wang2020Learning,hancock2018training} that parse explanations to logic forms, our approach does not require task-specific semantic parsers and matching models, making it task-agnostic and applicable to various natural language understanding tasks with minimal additional efforts.

%% file: experiments.tex
\section{Experiments}

\subsection{Datasets}

\begin{table} [!tb]
	\begin{center}
	    \scalebox{0.83}{
		\begin{tabular}{lrrrrr}
		\cline{1-5}
		    \toprule
		    \textbf{Dataset} & \textbf{\# Explanations} & \textbf{\# Train (Supervised)} & \textbf{\# Train (Semi-supervised)} & \textbf{\# Dev} & \textbf{\# Test} \\ 
		    \midrule
		    SemEval~\citep{hendrickx2010semeval} & 203 & 7,016 & 1,210  & 800 & 2,715 \\ 
		    TACRED~\citep{zhang2017tacred} & 139 & 68,006 & 2,751 & 22,531 & 15,509 \\ 
		    Laptop & 70 & 1,806 &135 & 462 & 638  \\ 
		    Restaurant & 75 & 2,830 & 107  & 720 & 1,120 \\ 
		    \bottomrule
	    \end{tabular}}
	\end{center}
	\caption{\textbf{Statistics of datasets.} We present the size of train/dev/test sets for 4 datasets in both supervised and semi-supervised settings. Moreover, \# Exp means the size of initial explanation sets. }
    \label{tab::data}
\end{table}

We conduct experiments on two tasks: relation extraction (RE) and aspect-based sentiment classification (ASC). For relation extraction we choose two datasets, TACRED~\cite{zhang2017position} and SemEval~\cite{hendrickx2010semeval} in our experiments. We use two customer review datasets, Restaurant and Laptop, which are part of SemEval 2014 Task 4~\cite{pontiki-etal-2014-semeval} for the aspect-based sentiment classification task. We use the human-annotated explanations collected in~\cite{Wang2020Learning} for training our explanation-based models. 

\subsection{Experimental Settings}
We conduct experiments in both the \textbf{supervised setting} where we have access to all labeled examples in the dataset and the \textbf{semi-supervised setting} where we only use a small fraction of labeled examples and considering the rest labeled examples in the original dataset as unlabeled examples by ignoring their labels. In both settings, only a few human-annotated NL explanations are available.
The number of explanations, labeled data used in supervised/unsupervised setting, and the statistics of the datasets are presented in Table \ref{tab::data}. 

We employ BERT-base and UniLM-base as the backbone of our prediction model and explanation generation model, respectively. We select batch size over $\{32, 64\}$ and learning rate over $\{\text{1e-5, 2e-5, 3e-5}\}$. The number of retrieved explanations is set to 10 for all tasks. We train the prediction model for 3 epochs and the generation model for 5 epochs in each EM iteration. We use Adam optimizers and early stopping with the best validation F1-score. 

\subsection{Compared Methods}

In the \textbf{supervised setting}, we compare \abbr with the BERT-base baseline that directly fine-tunes the pre-trained BERT-base model on the target datasets. To show the importance of modeling the interactions between the explanation generation model and the explanation-augmented prediction model, we also compare with a variant of our model, which only trains the explanation-augmented prediction module with all the explanations generated from the prior distribution, denoted as {ELV (M-step only)}).  
We also compare with some state-of-the-art algorithms on the RE and SA tasks.

\begin{table*}
\begin{floatrow}
\capbtabbox{
\resizebox{0.45\textwidth}{!}{
 \begin{tabular}{lcc}
 \toprule
 Method & TACRED & SemEval\\
 \midrule
 BERT$_{\text{EM}}$\cite{baldini-soares-etal-2019-matching} & 66.3 & 76.9 \\
 BERT$_{\text{EM+MTB}}$\cite{baldini-soares-etal-2019-matching} & \bf 67.1 & 77.5 \\
 BERT-large & 66.4 & 78.8 \\
 \midrule
 BERT-base &64.7 & 78.3  \\
  \midrule
 ELV (M-step only) & 65.4 & 80.2   \\
 \abbr (ours) & 65.9 & \bf 80.7 \\
 \bottomrule
 \end{tabular}
}}{
 \caption{Results (Micro-F1) on Relation Extraction datasets in supervised setting.}
 \label{tab:supre}
}
\capbtabbox{
\resizebox{0.45\textwidth}{!}{
 \begin{tabular}{lcc}
 \toprule
 Method & Restaurant & Laptop \\
 \midrule
 ASGCN~\cite{zhang2019aspect} & 72.2  &  71.1\\
  BERT-PT~\cite{xu2019bert} & 77.0  &  75.1\\
  BERT-SPC~\cite{song2019attentional} & 77.0 & 75.0 \\
\midrule
 BERT-base  & 75.4 & 72.4 \\
 \midrule
 ELV (M-step only) & 76.2 & 74.1 \\
 \abbr (ours) & \bf 77.8 & \bf 75.2 \\
 \bottomrule
 \end{tabular}
}}{
 \caption{Results (Macro-F1) on ASC datasets in supervised setting.}
 \label{tab:supasc}
}
\end{floatrow}
\end{table*}

In the \textbf{semi-supervised setting}, we compare ELV-EST against several competitive semi-supervised text classification methods including {Pseudo-Labeling}~\cite{lee2013pseudo}, {Self-Training}~\cite{rosenberg2005semi}, and {Data Programming}~\cite{hancock2018training} which incorporates NL explanations to perform semi-supervised text classification. Note that all compared model variants incorporate BERT-base as the backbone model.

\subsection{Experimental Results}

\textbf{Results on supervised setting. } We first present the results in the supervised setting in Table \ref{tab:supre} and \ref{tab:supasc}.  \abbr significantly outperforms the strong BERT baseline in all four datasets, demonstrating the effectiveness of exploiting NL explanations as additional information for natural language understanding. \abbr also consistently outperforms the ELV (M-step only), showing that \abbr's variational EM training effectively models the interactions between explanation and prediction. Also, the performance of \abbr compares favorably against several competitive recent studies focusing on RE and ASC respectively, further demonstrating the effectiveness of \abbr.

\begin{table*}
\begin{floatrow}
\capbtabbox{
\resizebox{0.45\textwidth}{!}{
 \begin{tabular}{lcc}
 \toprule
 Method & TACRED & SemEval \\
 \midrule
 BERT-base  & 25.1 & 49.3 \\
 \midrule
 Pseudo-Labeling~\cite{lee2013pseudo} & 28.6 & 50.2 \\
 Self-Training~\cite{rosenberg2005semi} & 36.9 & 59.5 \\
 Data Programming~\cite{hancock2018training} & 25.8 & 47.9 \\
 \midrule
 ELV-EST (ours) & \bf 42.5 & \bf 66.4\\
 \bottomrule
 \end{tabular}
}}{
 \caption{Results (Micro-F1) on Relation Extraction datasets in semi-supervised setting.}
 \label{tab:lowre}
}\vspace{-1mm}
\capbtabbox{
\resizebox{0.45\textwidth}{!}{
 \begin{tabular}{lcc}
 \toprule
 Method & Restaurant & Laptop \\
 \midrule
   BERT-base  &  32.2 & 34.6 \\
\midrule
 Pseudo-Labeling~\cite{lee2013pseudo} & 42.5 & 38.2   \\
 Self-Training~\cite{rosenberg2005semi} & 47.2 & 42.3 \\
 Data Programming~\cite{hancock2018training} & 38.2 & 36.3  \\
 \midrule
 ELV-EST (ours) & \bf 59.5 & \bf 63.6\\
 \bottomrule
 \end{tabular}
}}{
 \caption{Results (Macro-F1) on ASC datasets in semi-supervised setting.}
 \label{tab:lowasc}
}
\end{floatrow}
\end{table*}

\textbf{Results on semi-supervised setting. } The results in the semi-supervised setting are presented in Table \ref{tab:lowre} and \ref{tab:lowasc}. In the semi-supervised scenario, ELV-EST method significantly outperforms various semi-supervised text classification methods, as well as the data programming approach. The latter uses pre-defined rules to parse the NL explanations into logic forms and match unlabeled examples, on all four datasets. The improvement upon the BERT-base + self-training baseline is around 7 points for RE datasets and over 15 points for ASC datasets in terms of F1 score. This demonstrates the effectiveness of ELV-EST in the semi-supervised setting. 

\begin{table*}
\begin{floatrow}
\capbtabbox{
\resizebox{0.4\textwidth}{!}{
\begin{tabular}{lccc}
\toprule
Model & Inf. & Corr. & Cons. \\
\midrule
Seq2Seq & 2.43 & 3.27 & 2.68  \\
Transformer & 2.35 & 3.12 & 2.62  \\
UniLM & 3.48 & 3.94 & 3.14  \\
\midrule
ELV (ours) & \bf 3.87 & \bf 4.20 & \bf 3.51 \\
\bottomrule
\end{tabular}
}}
{\caption{\label{tab:humaneval} Human evaluation results. The scores scale from 1 to 5 (the larger, the better). The inner-rater agreement measured by Kappa score is 0.51.}}
\capbtabbox{
\resizebox{0.45\textwidth}{!}{
 \begin{tabular}{lcc}
 \toprule
 Method & Restaurant & Laptop \\
 \midrule
   BERT-base  & 75.4  & 72.4 \\
\midrule
 w. 80\% Rand Word        & 73.2 & 70.9 \\
 Orig + Rand Exp & 76.9 & 74.0  \\ \hline
    ELV (ours) & \bf 77.8 & \bf 75.2 \\
 \bottomrule
 \end{tabular}
}}{
 \caption{Results on ASC datasets with explanations with words randomly corrupted (80\%). Orig + Rand Exp is the 1:1 mix of original and randomly corrupted explanations. 
}
 \label{tab:randomexp}
}
\end{floatrow}
\end{table*}

\textbf{Results on explanation generation. } We further evaluate the quality of the explanation generation model with human evaluation. We invite 5 graduate students with enough English proficiency to score the explanations generated on the test set with input sentences and the labels predicted by the explanation-augmented prediction module\footnote{The prediction module is jointly trained with the explanation generation module}. The annotation scenarios include the explanantions' informativeness (Info.), correctness (Corr.), and consistency (Cons.) with respect to the model prediction. 
The inner-rater agreement is at 0.51 Kappa score. The details of human evaluation and examples of generated explanations are presented in the Appendix due to space constraints. 

For comparison, we include a fine-tuned UniLM model with annotated NL explanations, as well as two baselines trained from scratch using annotated NL explanations, one with a vanilla transformer model and the other an attention-based LSTM seq2seq model.
The results are in Table \ref{tab:humaneval}. The explanations generated by our \abbr framework are substantially better than those generated by the fine-tuned UniLM model. \abbr  generates better NL explanations that are relevant to the model's decision-making process, because it models the interactions of the explanation generation model and the prediction model. 

\subsection{Analysis}

\begin{figure}[h]%
    \centering
    {{\includegraphics[width=6cm]{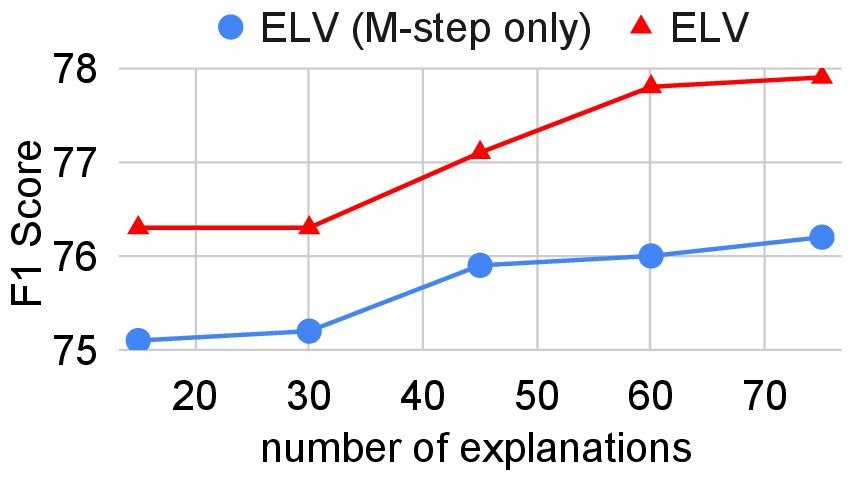} }}
    {{\includegraphics[width=6cm]{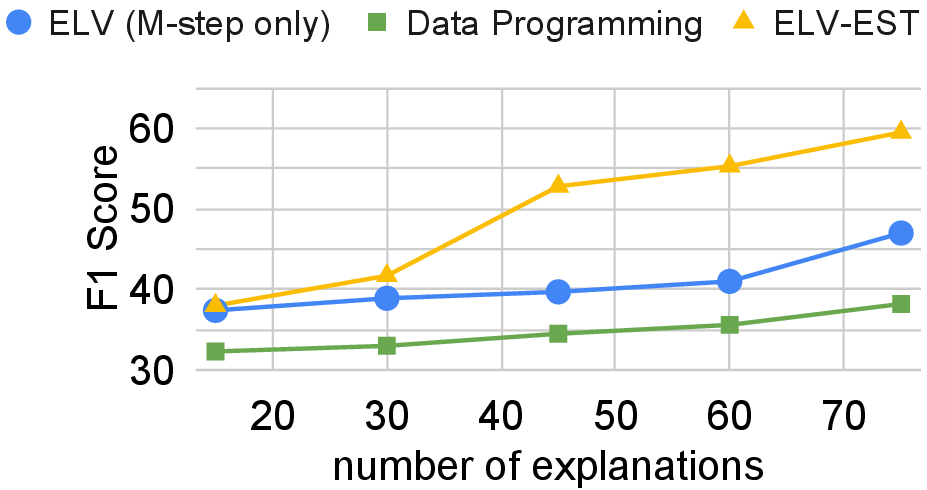} }}
    \caption{Performance with different number of explanations. We compare our method with baseline(s) in both supervised setting (left) and semi-supervised setting (right). }%
    \label{fig:exp_analysis}%
\end{figure}

\begin{figure}[h]%
    \centering
    {{\includegraphics[width=6cm]{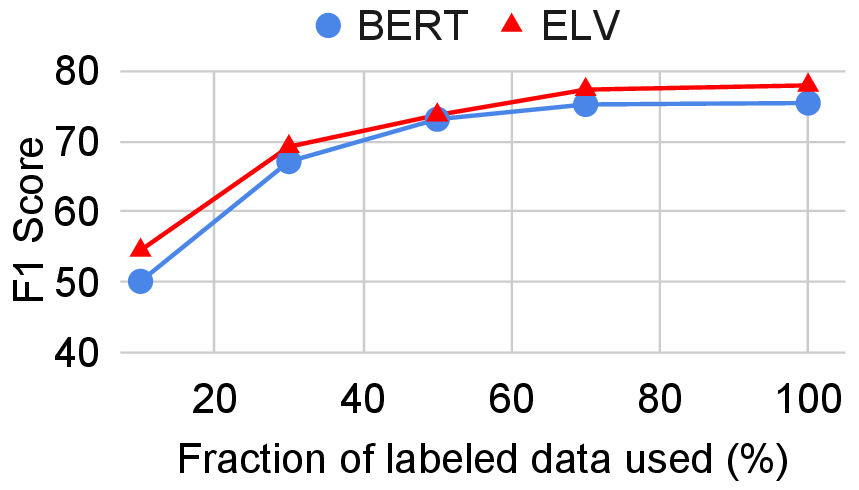} }}
    \ 
    {{\includegraphics[width=6cm]{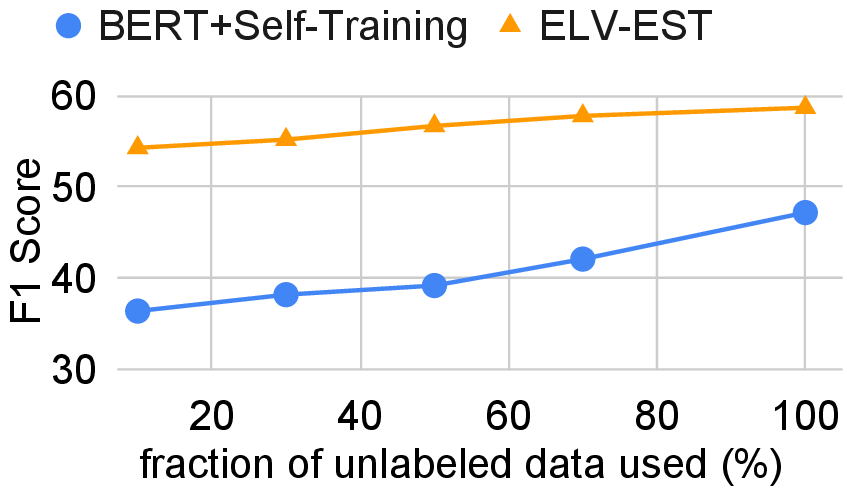} }}
    \caption{Performance with different number of labeled or unlabeled data in supervised setting (left) and semi-supervised setting (right) respectively. }
    \label{fig:label_analysis}%
\end{figure}

\textbf{Performance with corrupted explanations.} We first investigate the model performance w.r.t. the quality of the retrieved explanations. We compare with corrupted explanations which randomly replace 80\% of the words in the original explanations, results shown in Table~\ref{tab:randomexp}. The performance with corrupted explanations significantly decreases as expected. The high-quality explanations help the model better generalize while the random ones may confuse the model.

\textbf{Performance with different numbers of explanations.} 
We then investigate the performances with different amounts of explanations. As illustrated in Figure \ref{fig:exp_analysis} (left), with as few as 15 annotated explanations, \abbr significantly outperforms its counterpart trained without the variational EM framework in the supervised setting. 
The performance of \abbr continues to improve with more explanations but the performance of ELV (M-step only) starts to saturate with 45 explanations, showing the importance of modeling the interactions between the explanation generation model and explanation-augmented prediction model. Similar results are observed in the semi-supervised learning setting in Figure \ref{fig:exp_analysis} (right). 

\textbf{Performance with different numbers of labeled/unlabeled data.} We also investigate the performance of different models with different proportions of training data. From Figure \ref{fig:label_analysis} (left), we can see that \abbr consistently outperforms the BERT baseline with different amounts of labeled data. Especially, the improvement is the most significant when only 10\% of labeled data is used. This is because human explanations provide additional supervision and can serve as implicit logic rules to help generalization. In the semi-supervised learning setting (Figure \ref{fig:label_analysis}, right), ELV-EST outperforms traditional self-training methods by a large margin, especially with fewer unlabeled data, further confirming the improved generalization ability from explanations.

%% file: nips_2020.bbl
\begin{thebibliography}{40}
\providecommand{\natexlab}[1]{#1}
\providecommand{\url}[1]{\texttt{#1}}
\expandafter\ifx\csname urlstyle\endcsname\relax
  \providecommand{\doi}[1]{doi: #1}\else
  \providecommand{\doi}{doi: \begingroup \urlstyle{rm}\Url}\fi

\bibitem[Hancock et~al.(2018{\natexlab{a}})Hancock, Varma, Wang, Bringmann,
  Liang, and R{\'e}]{hancock-etal-2018-training}
Braden Hancock, Paroma Varma, Stephanie Wang, Martin Bringmann, Percy Liang,
  and Christopher R{\'e}.
\newblock Training classifiers with natural language explanations.
\newblock In \emph{Proceedings of the 56th Annual Meeting of the Association
  for Computational Linguistics (Volume 1: Long Papers)}, pages 1884--1895,
  Melbourne, Australia, July 2018{\natexlab{a}}. Association for Computational
  Linguistics.
\newblock \doi{10.18653/v1/P18-1175}.
\newblock URL \url{https://www.aclweb.org/anthology/P18-1175}.

\bibitem[Rajani et~al.(2019{\natexlab{a}})Rajani, McCann, Xiong, and
  Socher]{rajani-etal-2019-explain}
Nazneen~Fatema Rajani, Bryan McCann, Caiming Xiong, and Richard Socher.
\newblock Explain yourself! leveraging language models for commonsense
  reasoning.
\newblock In \emph{Proceedings of the 57th Annual Meeting of the Association
  for Computational Linguistics}, pages 4932--4942, Florence, Italy, July
  2019{\natexlab{a}}. Association for Computational Linguistics.
\newblock \doi{10.18653/v1/P19-1487}.
\newblock URL \url{https://www.aclweb.org/anthology/P19-1487}.

\bibitem[Camburu et~al.(2018)Camburu, Rockt{\"a}schel, Lukasiewicz, and
  Blunsom]{camburu2018snli}
Oana-Maria Camburu, Tim Rockt{\"a}schel, Thomas Lukasiewicz, and Phil Blunsom.
\newblock e-snli: Natural language inference with natural language
  explanations.
\newblock In \emph{Advances in Neural Information Processing Systems}, pages
  9539--9549, 2018.

\bibitem[Rajani et~al.(2019{\natexlab{b}})Rajani, McCann, Xiong, and
  Socher]{rajani2019explain}
Nazneen~Fatema Rajani, Bryan McCann, Caiming Xiong, and Richard Socher.
\newblock Explain yourself! leveraging language models for commonsense
  reasoning.
\newblock \emph{arXiv preprint arXiv:1906.02361}, 2019{\natexlab{b}}.

\bibitem[Srivastava et~al.(2017{\natexlab{a}})Srivastava, Labutov, and
  Mitchell]{srivastava-etal-2017-joint}
Shashank Srivastava, Igor Labutov, and Tom Mitchell.
\newblock Joint concept learning and semantic parsing from natural language
  explanations.
\newblock In \emph{Proceedings of the 2017 Conference on Empirical Methods in
  Natural Language Processing}, pages 1527--1536, Copenhagen, Denmark,
  September 2017{\natexlab{a}}. Association for Computational Linguistics.
\newblock \doi{10.18653/v1/D17-1161}.
\newblock URL \url{https://www.aclweb.org/anthology/D17-1161}.

\bibitem[Zhou et~al.(2020{\natexlab{a}})Zhou, Lin, Lin, Wang, Du, Neves, and
  Ren]{zhou2020nero}
Wenxuan Zhou, Hongtao Lin, Bill~Yuchen Lin, Ziqi Wang, Junyi Du, Leonardo
  Neves, and Xiang Ren.
\newblock Nero: A neural rule grounding framework for label-efficient relation
  extraction.
\newblock In \emph{Proceedings of The Web Conference 2020}, WWW ’20, page
  2166–2176, New York, NY, USA, 2020{\natexlab{a}}. Association for Computing
  Machinery.
\newblock ISBN 9781450370233.
\newblock \doi{10.1145/3366423.3380282}.
\newblock URL \url{https://doi.org/10.1145/3366423.3380282}.

\bibitem[Wang et~al.(2020)Wang, Qin, Zhou, Yan, Ye, Neves, Liu, and
  Ren]{Wang2020Learning}
Ziqi Wang, Yujia Qin, Wenxuan Zhou, Jun Yan, Qinyuan Ye, Leonardo Neves,
  Zhiyuan Liu, and Xiang Ren.
\newblock Learning from explanations with neural execution tree.
\newblock In \emph{International Conference on Learning Representations}, 2020.
\newblock URL \url{https://openreview.net/forum?id=rJlUt0EYwS}.

\bibitem[Palmer et~al.(2006)Palmer, Kreutz-Delgado, Rao, and
  Wipf]{palmer2006variational}
Jason Palmer, Kenneth Kreutz-Delgado, Bhaskar~D Rao, and David~P Wipf.
\newblock Variational em algorithms for non-gaussian latent variable models.
\newblock In \emph{Advances in neural information processing systems}, pages
  1059--1066, 2006.

\bibitem[Srivastava et~al.(2017{\natexlab{b}})Srivastava, Labutov, and
  Mitchell]{srivastava2017joint}
Shashank Srivastava, Igor Labutov, and Tom Mitchell.
\newblock Joint concept learning and semantic parsing from natural language
  explanations.
\newblock In \emph{Proceedings of the 2017 conference on empirical methods in
  natural language processing}, pages 1527--1536, 2017{\natexlab{b}}.

\bibitem[Murty et~al.(2020)Murty, Koh, and Liang]{murty2020expbert}
Shikhar Murty, Pang~Wei Koh, and Percy Liang.
\newblock Expbert: Representation engineering with natural language
  explanations.
\newblock \emph{arXiv preprint arXiv:2005.01932}, 2020.

\bibitem[Goldwasser and Roth(2014)]{goldwasser2014learning}
Dan Goldwasser and Dan Roth.
\newblock Learning from natural instructions.
\newblock \emph{Machine learning}, 94\penalty0 (2):\penalty0 205--232, 2014.

\bibitem[Fidler et~al.(2017)]{fidler2017teaching}
Sanja Fidler et~al.
\newblock Teaching machines to describe images with natural language feedback.
\newblock In \emph{Advances in Neural Information Processing Systems}, pages
  5068--5078, 2017.

\bibitem[Hancock et~al.(2018{\natexlab{b}})Hancock, Bringmann, Varma, Liang,
  Wang, and R{\'e}]{hancock2018training}
Braden Hancock, Martin Bringmann, Paroma Varma, Percy Liang, Stephanie Wang,
  and Christopher R{\'e}.
\newblock Training classifiers with natural language explanations.
\newblock In \emph{Proceedings of the conference. Association for Computational
  Linguistics. Meeting}, volume 2018, page 1884. NIH Public Access,
  2018{\natexlab{b}}.

\bibitem[Kingma and Welling(2013)]{kingma2013auto}
Diederik~P Kingma and Max Welling.
\newblock Auto-encoding variational bayes.
\newblock \emph{arXiv preprint arXiv:1312.6114}, 2013.

\bibitem[Bouraoui et~al.(2019)Bouraoui, Camacho-Collados, and
  Schockaert]{bouraoui2019inducing}
Zied Bouraoui, Jose Camacho-Collados, and Steven Schockaert.
\newblock Inducing relational knowledge from bert.
\newblock \emph{arXiv preprint arXiv:1911.12753}, 2019.

\bibitem[Petroni et~al.(2019)Petroni, Rockt{\"a}schel, Lewis, Bakhtin, Wu,
  Miller, and Riedel]{petroni2019language}
Fabio Petroni, Tim Rockt{\"a}schel, Patrick Lewis, Anton Bakhtin, Yuxiang Wu,
  Alexander~H Miller, and Sebastian Riedel.
\newblock Language models as knowledge bases?
\newblock \emph{arXiv preprint arXiv:1909.01066}, 2019.

\bibitem[Roberts et~al.(2020)Roberts, Raffel, and Shazeer]{roberts2020much}
Adam Roberts, Colin Raffel, and Noam Shazeer.
\newblock How much knowledge can you pack into the parameters of a language
  model?
\newblock \emph{arXiv preprint arXiv:2002.08910}, 2020.

\bibitem[Dong et~al.(2019)Dong, Yang, Wang, Wei, Liu, Wang, Gao, Zhou, and
  Hon]{dong2019unified}
Li~Dong, Nan Yang, Wenhui Wang, Furu Wei, Xiaodong Liu, Yu~Wang, Jianfeng Gao,
  Ming Zhou, and Hsiao-Wuen Hon.
\newblock Unified language model pre-training for natural language
  understanding and generation.
\newblock In \emph{Advances in Neural Information Processing Systems}, pages
  13042--13054, 2019.

\bibitem[Reimers and Gurevych(2019)]{reimers-gurevych-2019-sentence}
Nils Reimers and Iryna Gurevych.
\newblock Sentence-{BERT}: Sentence embeddings using {S}iamese {BERT}-networks.
\newblock In \emph{Proceedings of the 2019 Conference on Empirical Methods in
  Natural Language Processing and the 9th International Joint Conference on
  Natural Language Processing (EMNLP-IJCNLP)}, pages 3982--3992, Hong Kong,
  China, November 2019. Association for Computational Linguistics.
\newblock \doi{10.18653/v1/D19-1410}.
\newblock URL \url{https://www.aclweb.org/anthology/D19-1410}.

\bibitem[Devlin et~al.(2019)Devlin, Chang, Lee, and
  Toutanova]{devlin-etal-2019-bert}
Jacob Devlin, Ming-Wei Chang, Kenton Lee, and Kristina Toutanova.
\newblock {BERT}: Pre-training of deep bidirectional transformers for language
  understanding.
\newblock In \emph{Proceedings of the 2019 Conference of the North {A}merican
  Chapter of the Association for Computational Linguistics: Human Language
  Technologies, Volume 1 (Long and Short Papers)}, pages 4171--4186,
  Minneapolis, Minnesota, June 2019. Association for Computational Linguistics.
\newblock \doi{10.18653/v1/N19-1423}.
\newblock URL \url{https://www.aclweb.org/anthology/N19-1423}.

\bibitem[Hendrickx et~al.(2010)Hendrickx, Kim, Kozareva, Nakov, S{\'e}aghdha,
  Pad{\'o}, Pennacchiotti, Romano, and Szpakowicz]{hendrickx2010semeval}
Iris Hendrickx, Su~Nam Kim, Zornitsa Kozareva, Preslav Nakov, Diarmuid~O
  S{\'e}aghdha, Sebastian Pad{\'o}, Marco Pennacchiotti, Lorenza Romano, and
  Stan Szpakowicz.
\newblock Semeval-2010 task 8: Multi-way classification of semantic relations
  between pairs of nominals.
\newblock In \emph{Proceedings of the 5th International Workshop on Semantic
  Evaluation}, pages 33--38. Association for Computational Linguistics, 2010.

\bibitem[Zhang et~al.(2017{\natexlab{a}})Zhang, Zhong, Chen, Angeli, and
  Manning]{zhang2017tacred}
Yuhao Zhang, Victor Zhong, Danqi Chen, Gabor Angeli, and Christopher~D.
  Manning.
\newblock Position-aware attention and supervised data improve slot filling.
\newblock In \emph{Proceedings of the 2017 Conference on Empirical Methods in
  Natural Language Processing (EMNLP 2017)}, pages 35--45, 2017{\natexlab{a}}.
\newblock URL \url{https://nlp.stanford.edu/pubs/zhang2017tacred.pdf}.

\bibitem[Zhang et~al.(2017{\natexlab{b}})Zhang, Zhong, Chen, Angeli, and
  Manning]{zhang2017position}
Yuhao Zhang, Victor Zhong, Danqi Chen, Gabor Angeli, and Christopher~D Manning.
\newblock Position-aware attention and supervised data improve slot filling.
\newblock In \emph{Proceedings of the 2017 Conference on Empirical Methods in
  Natural Language Processing}, pages 35--45, 2017{\natexlab{b}}.

\bibitem[Pontiki et~al.(2014)Pontiki, Galanis, Pavlopoulos, Papageorgiou,
  Androutsopoulos, and Manandhar]{pontiki-etal-2014-semeval}
Maria Pontiki, Dimitris Galanis, John Pavlopoulos, Harris Papageorgiou, Ion
  Androutsopoulos, and Suresh Manandhar.
\newblock {S}em{E}val-2014 task 4: Aspect based sentiment analysis.
\newblock In \emph{Proceedings of the 8th International Workshop on Semantic
  Evaluation ({S}em{E}val 2014)}, pages 27--35, Dublin, Ireland, August 2014.
  Association for Computational Linguistics.
\newblock \doi{10.3115/v1/S14-2004}.
\newblock URL \url{https://www.aclweb.org/anthology/S14-2004}.

\bibitem[Baldini~Soares et~al.(2019)Baldini~Soares, FitzGerald, Ling, and
  Kwiatkowski]{baldini-soares-etal-2019-matching}
Livio Baldini~Soares, Nicholas FitzGerald, Jeffrey Ling, and Tom Kwiatkowski.
\newblock Matching the blanks: Distributional similarity for relation learning.
\newblock In \emph{Proceedings of the 57th Annual Meeting of the Association
  for Computational Linguistics}, pages 2895--2905, Florence, Italy, July 2019.
  Association for Computational Linguistics.
\newblock \doi{10.18653/v1/P19-1279}.
\newblock URL \url{https://www.aclweb.org/anthology/P19-1279}.

\bibitem[Zhang et~al.(2019)Zhang, Li, and Song]{zhang2019aspect}
Chen Zhang, Qiuchi Li, and Dawei Song.
\newblock Aspect-based sentiment classification with aspect-specific graph
  convolutional networks.
\newblock \emph{arXiv preprint arXiv:1909.03477}, 2019.

\bibitem[Xu et~al.(2019)Xu, Liu, Shu, and Yu]{xu2019bert}
Hu~Xu, Bing Liu, Lei Shu, and Philip~S Yu.
\newblock Bert post-training for review reading comprehension and aspect-based
  sentiment analysis.
\newblock \emph{arXiv preprint arXiv:1904.02232}, 2019.

\bibitem[Song et~al.(2019)Song, Wang, Jiang, Liu, and Rao]{song2019attentional}
Youwei Song, Jiahai Wang, Tao Jiang, Zhiyue Liu, and Yanghui Rao.
\newblock Attentional encoder network for targeted sentiment classification.
\newblock \emph{arXiv preprint arXiv:1902.09314}, 2019.

\bibitem[Lee(2013)]{lee2013pseudo}
Dong-Hyun Lee.
\newblock Pseudo-label: The simple and efficient semi-supervised learning
  method for deep neural networks.
\newblock In \emph{Workshop on challenges in representation learning, ICML},
  volume~3, page~2, 2013.

\bibitem[Rosenberg et~al.(2005)Rosenberg, Hebert, and
  Schneiderman]{rosenberg2005semi}
Chuck Rosenberg, Martial Hebert, and Henry Schneiderman.
\newblock Semi-supervised self-training of object detection models.
\newblock \emph{WACV/MOTION}, 2, 2005.

\bibitem[Yang et~al.(2019)Yang, Dai, Yang, Carbonell, Salakhutdinov, and
  Le]{yang2019xlnet}
Zhilin Yang, Zihang Dai, Yiming Yang, Jaime Carbonell, Russ~R Salakhutdinov,
  and Quoc~V Le.
\newblock Xlnet: Generalized autoregressive pretraining for language
  understanding.
\newblock In \emph{Advances in neural information processing systems}, pages
  5753--5763, 2019.

\bibitem[Liu et~al.(2019)Liu, Ott, Goyal, Du, Joshi, Chen, Levy, Lewis,
  Zettlemoyer, and Stoyanov]{liu2019roberta}
Yinhan Liu, Myle Ott, Naman Goyal, Jingfei Du, Mandar Joshi, Danqi Chen, Omer
  Levy, Mike Lewis, Luke Zettlemoyer, and Veselin Stoyanov.
\newblock Roberta: A robustly optimized bert pretraining approach.
\newblock \emph{arXiv preprint arXiv:1907.11692}, 2019.

\bibitem[Lan et~al.(2019)Lan, Chen, Goodman, Gimpel, Sharma, and
  Soricut]{lan2019albert}
Zhenzhong Lan, Mingda Chen, Sebastian Goodman, Kevin Gimpel, Piyush Sharma, and
  Radu Soricut.
\newblock Albert: A lite bert for self-supervised learning of language
  representations.
\newblock \emph{arXiv preprint arXiv:1909.11942}, 2019.

\bibitem[Clark et~al.(2020)Clark, Luong, Le, and Manning]{clark2020electra}
Kevin Clark, Minh-Thang Luong, Quoc~V Le, and Christopher~D Manning.
\newblock Electra: Pre-training text encoders as discriminators rather than
  generators.
\newblock \emph{arXiv preprint arXiv:2003.10555}, 2020.

\bibitem[Sanh et~al.(2019)Sanh, Debut, Chaumond, and Wolf]{sanh2019distilbert}
Victor Sanh, Lysandre Debut, Julien Chaumond, and Thomas Wolf.
\newblock Distilbert, a distilled version of bert: smaller, faster, cheaper and
  lighter.
\newblock \emph{arXiv preprint arXiv:1910.01108}, 2019.

\bibitem[Xu et~al.(2020)Xu, Zhou, Ge, Wei, and Zhou]{xu2020bert}
Canwen Xu, Wangchunshu Zhou, Tao Ge, Furu Wei, and Ming Zhou.
\newblock Bert-of-theseus: Compressing bert by progressive module replacing.
\newblock \emph{arXiv preprint arXiv:2002.02925}, 2020.

\bibitem[Xin et~al.(2020)Xin, Tang, Lee, Yu, and Lin]{xin2020deebert}
Ji~Xin, Raphael Tang, Jaejun Lee, Yaoliang Yu, and Jimmy Lin.
\newblock Deebert: Dynamic early exiting for accelerating bert inference.
\newblock \emph{arXiv preprint arXiv:2004.12993}, 2020.

\bibitem[Zhou et~al.(2020{\natexlab{b}})Zhou, Xu, Ge, McAuley, Xu, and
  Wei]{zhou2020bert}
Wangchunshu Zhou, Canwen Xu, Tao Ge, Julian McAuley, Ke~Xu, and Furu Wei.
\newblock Bert loses patience: Fast and robust inference with early exit.
\newblock \emph{arXiv preprint arXiv:2006.04152}, 2020{\natexlab{b}}.

\bibitem[Zhao et~al.(2017)Zhao, Wang, Yatskar, Ordonez, and Chang]{zhao2017men}
Jieyu Zhao, Tianlu Wang, Mark Yatskar, Vicente Ordonez, and Kai-Wei Chang.
\newblock Men also like shopping: Reducing gender bias amplification using
  corpus-level constraints.
\newblock \emph{arXiv preprint arXiv:1707.09457}, 2017.

\bibitem[Costa-juss{\`a}(2019)]{costa2019analysis}
Marta~R Costa-juss{\`a}.
\newblock An analysis of gender bias studies in natural language processing.
\newblock \emph{Nature Machine Intelligence}, pages 1--2, 2019.

\end{thebibliography}
